\documentclass[%
    reprint,
    superscriptaddress,
    longbibliography,
    nofootinbib,
    amsmath,
    amssymb,
    aps,
    pra,
    floatfix,
]{revtex4-2}

\usepackage[normalem]{ulem}
\usepackage{xcolor}
\usepackage{graphicx}
\usepackage{dcolumn}
\usepackage{bm}
\usepackage[utf8]{inputenc}
\usepackage{acronym}
\usepackage{graphicx}
\usepackage{amsmath,amsfonts,amssymb}
\usepackage{braket}
\usepackage{tikz}
\usetikzlibrary{quantikz}
\usepackage{acronym}
\usepackage{xspace}
\usepackage{siunitx}
\usepackage{booktabs}
\usepackage[hidelinks]{hyperref}
\usepackage{nicefrac}       
\usepackage{microtype}      
\usepackage{mathtools}
\usepackage{listings}
\usepackage{todonotes} 
\usepackage{siunitx}
\usepackage{multirow}
\usepackage{tabularx}
\usepackage{float}

\newacro{LHC}{Large Hadron Collider}
\newacro{HEP}{High Energy Physics}
\newacro{QML}{Quantum Machine Learning}
\newacro{ML}{Machine Learning}
\newacro{SM}{Standard Model}
\newacro{BSM}{Beyond Standard Model}
\newacro{MC}{Monte Carlo}
\newacro{AD}{Anomaly Detection}
\newacro{RS}{Randall-Sundrum}
\newacro{AE}{autoencoder}
\newacro{QSVM}{Quantum Support Vector Machines}
\newacro{NISQ}{Noisy Intermediate Scale Quantum}
\newacro{TPR}{True Positive Rate}
\newacro{FPR}{False Positive Rate}

\begin{document}
\title{Guided Graph Compression for Quantum Graph Neural Networks}
\author{Mikel Casals}
\email{mikel.casals@estudiantat.upc.edu}
\affiliation{Universitat Politècnica de Catalunya, 08034 Barcelona, Spain}
\author{Vasilis Belis}
\affiliation{Institute for Particle Physics and Astrophysics, ETH Zurich, 8093 Zurich, Switzerland}
\author{Elias F. Combarro}
\affiliation{Computer Science Department, University of Oviedo, 33003 Oviedo, Spain}
\author{Eduard Alarcón}
\affiliation{Universitat Politècnica de Catalunya, 08034 Barcelona, Spain}
\author{Sofia Vallecorsa}
\affiliation{European Organization for Nuclear Research (CERN), CH-1211 Geneva, Switzerland}    
\author{Michele Grossi}
\affiliation{European Organization for Nuclear Research (CERN), CH-1211 Geneva, Switzerland}

\date{\today}
\begin{abstract}
Graph Neural Networks (GNNs) are effective for processing graph-structured data but face challenges with large graphs due to high memory requirements and inefficient sparse matrix operations on GPUs. Quantum Computing (QC) offers a promising avenue to address these issues and inspires new algorithmic approaches. In particular, Quantum Graph Neural Networks (QGNNs) have been explored in recent literature. However, current quantum hardware limits the dimension of the data that can be effectively encoded. Existing approaches either simplify datasets manually or use artificial graph datasets. This work introduces the Guided Graph Compression (GGC) framework, which uses a graph autoencoder to reduce both the number of nodes and the dimensionality of node features. The compression is guided to enhance the performance of a downstream classification task, which can be applied either with a quantum or a classical classifier. The framework is evaluated on the Jet Tagging task, a classification problem of fundamental importance in high energy physics that involves distinguishing particle jets initiated by quarks from those by gluons. The GGC is compared against using the autoencoder as a standalone preprocessing step and against a baseline classical GNN classifier. Our numerical results demonstrate that GGC outperforms both alternatives, while also facilitating the testing of novel QGNN ansatzes on realistic datasets.
\end{abstract}
\maketitle

\section{Introduction}

A graph is a flexible data structure that can be used to represent many complex systems. Graph Neural Networks (GNNs) are a set of Machine Learning (ML) models that are powerful for analyzing graph-structured data. For instance, they have been used for social media recommendations \cite{fan2019graph}, drug-target interaction prediction \cite{veleiro2024gennius}, and particle tracking reconstruction \cite{ju2020graph} in high energy physics. However, classical GNNs suffer from limitations when handling large graphs, since the memory requirements are high \cite{chiang2019cluster} and they require sparse matrix operations, which are not efficient in classical GPUs \cite{abadal2021computing}. Moreover, GNNs are also limited in expressiveness by the Weisfeiler-Lehman test \cite{feng2022powerful}.

As a result, there is motivation to explore Quantum Computing (QC) approaches to overcome these limitations \cite{liao2024graph}. QC is an emerging computing paradigm that is known to surpass classical methods in terms of speed for some problems \cite{shor1999polynomial,grover1996fast,harrow2009quantum}. Furthermore, Quantum Machine Learning (QML) aims to enhance classical ML methods using quantum computers \cite{schuld2021machine}. Studies on the generalization and expressivity of QML models~\cite{barthe2024parameterized, jerbi2023quantum, perez2020data} highlight their potential and identify specific problems where QML outperforms classical methods~\cite{caro2022generalization,Liu2021rigorous, Muser:2023tos, huangQA2022, Huang2021}.

One of the most popular QML frameworks is the Variational Quantum Circuit (VQC) \cite{cerezo2021variational}. Within this framework, Quantum Neural Networks (QNNs) \cite{farhi2018classification} serve as the quantum analogs to classical Neural Networks (NNs). They consist of training a parameterized quantum circuit, usually referred to as the ansatz, by optimizing its parameters using a classical computer. General QNN ansatzes have already been used for a number of ML tasks~\cite{schuld2020circuit, abbas2021power, wan2017quantum, blance2021quantum, ngairangbam2022anomaly}. However, these types of general ansatzes suffer from barren plateaus \cite{larocca2024reviewbp}, where the gradients of the quantum circuit vanish. An approach to addressing this issue is to employ quantum circuit architectures with inductive biases tailored to the problem at hand, instead of relying on generic ansatzes~\cite{Nguyen:2022lww}. Quantum Graph Neural Networks (QGNNs) \cite{verdon2019quantum} emerged with this motivation, enabling quantum computers to process graph-structured data by embedding permutation-equivariant node operations and permutation-invariant graph readouts \cite{keriven2019universal}.


For non-graph data, autoencoders have been employed in quantum machine learning (QML) to reduce dataset dimensionality \cite{slabbert2024, belis2024quantum, mangini2022quantum}. Building on this, the Guided Quantum Compression (GQC) framework, introduced in \cite{belis2024guided}, enhances dimensionality reduction by jointly optimizing for data compression and classification performance in a complex High Energy Physics (HEP) task. This is achieved by combining the reconstruction loss of the autoencoder with the classification loss of a quantum classifier. Applied to the task of identifying Higgs boson production in proton collisions, the GQC framework outperforms approaches where the autoencoder is trained as a separate preprocessing step.

In this work, we generalize the GQC framework to graph-structured data by introducing the Guided Graph Compression (GGC) framework. Here, graph autoencoders are used to compress graphs both in terms of the number of nodes and the dimensionality of node features. As in the GQC, the reconstruction and classification loss functions are jointly optimized, guiding the compression process to improve downstream classification performance.

We apply this framework to the Jet Tagging task \cite{gallicchio2011quark}, which involves classifying jets---sets of particles represented as graphs---as originating from either quarks or gluons. To the best of our knowledge, GGC is the first architecture that combines the compression of graphs with a classification model, as a unified procedure in classical and quantum ML. Our results show that guided graph compression significantly outperforms the use of a graph autoencoder as a standalone preprocessing step, which is the standard approach used in the literature. 

\section{Models}
\begin{figure*}[thb]
    \centering
    \includegraphics[width=1\textwidth]{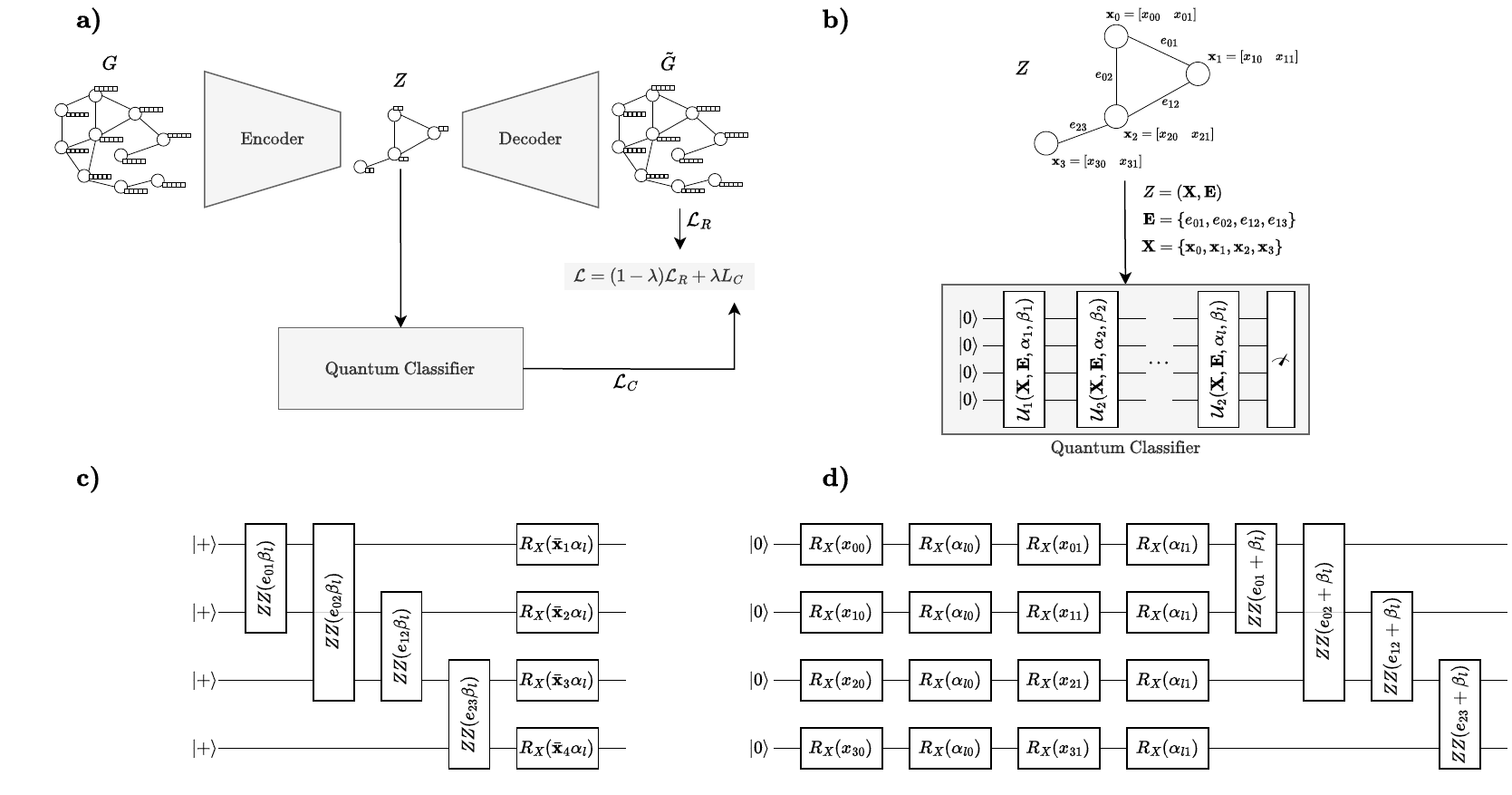}
    \caption{The Guided Graph Compression (GGC) framework. A high-level overview of the GGC architecture is shown in (a). The autoencoder receives an input graph $G$ and produces a latent compressed graph $Z$ via the Encoder. The Decoder receives $Z$ and aims to reconstruct the original graph $G$. The latent graph $Z$ is also fed to the Quantum Classifier, as shown in (b). The Quantum Classifier is formed of feature maps $U(\cdot)$ that encode the set of node features $\mathbf{X}$ and the set of edge weights $\mathbf{E}$, using a data re-uploading strategy. They also contain trainable parameters $\alpha_l$ and $\beta_l$. The two feature maps that are investigated in this work are (c) QGNN1 and (d) QGNN2.}
    \label{fig:ggc}
\end{figure*}

The Guided Graph Compression (GGC) framework consists of a graph autoencoder and a classifier, as illustrated in Figure \ref{fig:ggc}. In the following, we introduce its key components.

\subsection{Graph Autoencoders}
We investigate two graph autoencoders, introduced in~\cite{ge2021graph}, which construct a compressed latent representations of graphs in terms of number of nodes and node feature dimensionality. The loss function used for these models is the Mean Squared Error (MSE) between the original and reconstructed node features
\begin{equation} \label{eq:mse}
    \mathcal{L}_R = \frac{1}{M}\sum^M_{m=1}\left[x_m - \mathcal{D}\circ\mathcal{E}(x_m)\right]^2,    
\end{equation}
where $M$ is the number of graph data samples, $x_m$ represents the original node features of graph $m$, $\mathcal{E}$ is the encoder network and $\mathcal{D}$ is the decoder network. 

We introduce the two autoencoder models in the following subsections.

\subsubsection{Multi-Kernel Inductive Attention Graph model}
The Multi-Kernel Inductive Attention Graph Autoencoder (MIAGAE) has an encoder network and a decoder network. The encoder is trained to obtain the compressed graph and the decoder is trained to reconstruct the original graph using the latent representation.

The encoder can be divided into 2 sublayers:

\begin{itemize}
    \item Multi-Kernel Inductive Convolutional (MI-Conv) layer: based on the GraphSAGE architecture \cite{hamilton2017inductive}, this layer uses multiple kernels to learn different aspects of a feature. The features extracted by the $m-$th kernel at layer $k+1$ for node $i$ are obtained by
    \begin{equation} \label{eq:miagae_nodefeat}
\hat{f}^{i}_{k+1} = W_1^m f^{i}_{k} + W_2^m \cdot \texttt{mean}_{j \in \mathcal{N}(i)} f^{j}_{k},
    \end{equation}
    where $W_1^m$ and $W_2^m$ are trainable weight matrices, \texttt{mean} is the aggregation function and $\mathcal{N}(i)$ is the neighborhood of node $i$. The final node features of node $i$ at layer $k+1$ are obtained using \begin{equation}\label{eq:miagae_kernels}
        f^{i}_{k+1} = \texttt{Aggre}\left(\sigma(\hat{f}^{i}_{k+1,1}), \sigma(\hat{f}^{i}_{k+1,2}), ..., \sigma(\hat{f}^{i}_{k+1,m})\right),
    \end{equation}
    where $\texttt{Aggre}$ is generally any aggregation function and $\sigma$ is an activation function. In this study, the sum aggregation function and \texttt{ReLU} are used, respectively.

    \item Similarity Attention Graph Pooling (SimAGPool) layer: it downsamples the number of nodes in the graph. To keep the nodes that contain most of the graph's information, a score is calculated for all the nodes and the top $p$ percent of nodes are kept, where $p \in (0,1]$ is the pooling ratio. The score is named Representativeness and Contribution Score (RCS) and is given by
    \begin{equation}\label{eq:miagae_rcs}
        \text{RCS}_{i} = \sum_{j \in \mathcal{N}(i)} ({f^i}^T f^j) \quad \text{s.t.} \quad \exists e_{ij},
    \end{equation}
    
    where $f^i$ and $f^j$ are the features of node $i$ and $j$ respectively and $e_{ij}$ represents the edge between these nodes.
\end{itemize}

The decoder contains an Unpooling layer which is symmetric to the SimAGPool layer. It takes the output graph of the previous layer and the edge information of removed nodes to reconstruct the original graph. It has the same parameter settings as the MI-Conv layer. 

The graph autoencoder contains the following hyperparameters:

\begin{itemize}
    \item Depth: the number of layers in the encoder and decoder
    \item Shapes: dimension of the node features at each layer in the encoder. In the decoder, it is automatically reversed.
    \item Compression rate: ratio of compression for each layer in the encoder
    \item Kernels: number of kernels in the MI-Conv layer.
\end{itemize}

\subsubsection{Self-Attention Graph model}
The main difference between the SAG model and MIAGAE is the pooling layer, which is substituted by a Self-Attention Graph (SAG) pooling layer \cite{lee2019self}. SAG pooling uses a GNN to obtain self-attention scores for each node. The simplest mechanism uses graph convolution as its GNN and the scores are calculated with Equation \ref{eq:sag_pooling},

\begin{equation}\label{eq:sag_pooling}
    Z = \sigma \left( \tilde{D}^{-\frac{1}{2}} \tilde{A} \tilde{D}^{-\frac{1}{2}} X \Theta_{\text{att}} \right),
\end{equation}
where $\tilde{D}$ represents the degree matrix of the graph, $\tilde{A}$ is the adjacency matrix of the graph with self-connections, $X$ is the input feature matrix of the graph, $\Theta_{\text{att}}$ are trainable parameters and $\sigma$ represents the non-linear activation function; in this case, $\tanh$. The top $p$ percent nodes with highest scores are kept.

\subsection{Classifiers}
The GGC framework can be used both with classical and quantum classifiers. In this work, a classical baseline classifier and two quantum classifiers are used to evaluate its performance. The loss function used for all the classifiers is the Binary Cross Entropy (BCE),

\begin{equation}\label{eq:bce}
   \mathcal{L}_C = -\frac{1}{M} \sum_{m=1}^{M} \left[ y_m \log(\hat{y}_m) + (1 - y_m) \log(1 - \hat{y}_m) \right], 
\end{equation}
where $M$ is the number of graph data samples, $y_m$ is the true label of the $m$-th data sample and $\hat{y}_m$ is the corresponding predicted probability. That is, a continuous value between 0 and 1 representing the probability of the example belonging to the class according to the classifier.

\subsubsection{Classical}
A classical GNN-based classifier is used to set a baseline to which the quantum classifiers are compared to.

\begin{itemize}
    \item Classical GNN: a basic Graph Convolutional Neural Network (GCN) model that consists of a convolutional layer that aggregates information from neighboring nodes, a pooling layer that outputs a feature vector of the whole graph, and a fully connected neural network layer that gives the final prediction.
\end{itemize}

\subsubsection{Quantum}\label{sec:qgnn}
Two quantum classifiers found in the literature are used to evaluate the performance of GGC. Note that the purpose of this work is not to obtain the best quantum classifier possible. Further research is required to design better quantum classifiers. The choice of the following two classifiers has been made taking into account the simplicity of implementation and their performance on their respective works.

\begin{itemize}
    \item Quantum Graph Neural Network 1 (QGNN1): it is a trainable permutation equivariant embedding circuit that was introduced in \cite{skolik2023equivariant}. One layer of the quantum circuit can be seen in Figure \ref{fig:ggc}(c). Every node is embedded into a qubit and they are initialized in the $\ket+$ state. Then, $ZZ$ gates are applied to embed the edge weights of the graph $e_{ij}$ with a shared trainable parameter $\beta_l$. Lastly, $R_x$ gates are applied to each qubit, embedding each node feature $x_m$ with a shared trainable parameter $\alpha_l$. If the dimensionality of the node features is more than 1, the mean is taken. The trainable parameters $\alpha_l$ and $\beta_l$ are unique for each layer.


    \item Quantum Graph Neural Network 2 (QGNN2): it is a trainable permutation equivariant embedding circuit that was introduced in~\cite{yeequivariant}. An example of a layer of this QGNN can be seen in Figure \ref{fig:ggc}(d). In this case, every node is initialized in the $\ket0$ state. Then, $R_x$ gates are applied. $x_{mk}$ represents the $k$-th feature of node $m$, and $\alpha_{lk}$ represents a trainable parameter corresponding to the $k$-th feature at layer $l$. Lastly, $ZZ$ gates are applied to embed the edge weights $e_{ij}$ by summing them with a trainable parameter $\beta_l$.

\end{itemize}

Both introduced circuit layers are repeated $l$ times using a data re-uploading strategy \cite{Perez-Salinas:2019pjx}, as it can be seen in Figure \ref{fig:ggc}(b). 

The output of the circuits is the expected value of the $Z$ observable applied to the first qubit, which is interpreted as the probability that the input graph belongs to the positive class. Specifically, the output of the model is:
\begin{equation}\label{eq:measurement}   f_\vartheta(z)=\Braket{0|\mathcal{U}_\vartheta^\dagger(z)Z\,\mathcal{U}_\vartheta\left(z\right)|0},
\end{equation}
where $\mathcal{U}_\vartheta(z)$ is the quantum circuit, $z$ is the compressed graph and $\ket{0}=\ket{0}^{\otimes n}$ is the initial n-qubit state. We assume without loss of generality that $f_\vartheta(z) \in [-1,1]$. Therefore, the predicted label $\hat{y}$ from the quantum circuit is \begin{equation}\label{eq:sign}
    \hat{y} = \frac{\text{sign}\left[f_\vartheta(z) \right]+1}{2}.
\end{equation}

\section{Training Paradigms}

In this work, three different training paradigms are evaluated:
\begin{itemize}
    \item Uncompressed classical: it consists of training the classical GNN classifier with the raw data without compression.

    \item Not guided graph compression (two-step approach): it consists of training a graph autoencoder as an independent preprocessing step. Then, from the trained autoencoder, the latent graphs are obtained to train the classifiers, both classical and quantum.

    \item Guided graph compression: it consists of training a graph autoencoder and a classifier by combining both loss functions. The total loss function is
    \begin{equation}\label{eq:ggc}
        \mathcal{L}=(1-\lambda)\mathcal{L}_R + \lambda\mathcal{L}_C,
    \end{equation}
    where $\lambda$ is a hyperparameter that is used to vary the weight of the losses.
\end{itemize}

\section{Results}
To evaluate the performance of the GGC framework on a realistic dataset, we consider a complex classification task from High Energy Physics (HEP): Jet Tagging \cite{gallicchio2011quark}. Jet Tagging is a fundamental problem in HEP, as distinguishing whether jets---collimated sprays of particles that can be represented as graphs---originate from quarks or gluons plays a key role in characterizing events at particle colliders and in searching for new physics.

For this task, we use the Pythia8 Quark and Gluon Jets for Energy Flow dataset \cite{komiske2019energy}, which excludes charm and bottom quark jets. It contains 2 million jet samples, balanced between the two classes. Each jet consists of a variable number of particles, initially described by 4 features, which are augmented to 13 features following the approach in \cite{qu2022particle}. Jets are then represented as fully connected graphs, as in \cite{qu2020jet}. More details on dataset preprocessing are provided in Appendix \ref{appendix:feat_engineering}.


\begin{table}[thb]
    \centering
    \begin{tabular}{ll}
        \toprule
        & \textbf{Test ROC-AUC} \\
        \midrule
        \textbf{Uncompressed} \\
        \quad GNN & $0.8051 \pm 0.0041$ \\
        \cmidrule(lr){1-2} 
        \textbf{Not Guided Graph Compression} \\  
        \quad MIAGAE + GNN  & $0.5797 \pm 0.0049$ \\
        \quad MIAGAE + QGNN1  & $0.6007 \pm 0.0055$\\
        \quad MIAGAE + QGNN2  & $0.7811 \pm 0.0046$ \\
        \cmidrule(lr){1-2} 
        \textbf{Guided Graph Compression} \\
        \quad MIAGAE + GNN   & $\mathbf{0.8645 \pm 0.0033}$ \\
        \quad MIAGAE + QGNN1   & $0.8021 \pm 0.0048$ \\
        \quad MIAGAE + QGNN2   & $0.8406 \pm 0.0039$ \\
        \bottomrule
    \end{tabular}
    \caption{Summary of the test ROC-AUC results for all training paradigms using the MIAGAE as the graph autoencoder.}
    \label{tab:summary_miagae}
\end{table}
\begin{table} [thb]
    \centering
    \begin{tabular}{ll}
        \toprule
        & \textbf{Test ROC-AUC} \\
        \midrule
        \textbf{Uncompressed} \\
        \quad GNN & $0.8051 \pm 0.0041$ \\
        \cmidrule(lr){1-2} 
        \textbf{Not Guided Graph Compression} \\
        \quad SAG model + GNN  & $0.6741 \pm 0.0050$ \\
        \quad SAG model + QGNN1  & $0.7004 \pm 0.0050 $ \\
        \quad SAG model + QGNN2  & $0.7297 \pm 0.0050$ \\
        \cmidrule(lr){1-2} 
        \textbf{Guided Graph Compression} \\
        \quad SAG model + GNN   & $0.8697 \pm 0.0033$ \\
        \quad SAG model + QGNN1   & $0.7466 \pm 0.0038$ \\
        \quad SAG model + QGNN2   & $\mathbf{0.8721 \pm 0.0030}$ \\
        \bottomrule
    \end{tabular}
    \caption{Summary of the test ROC-AUC results for all training paradigms using the SAG model as the graph autoencoder.}
    \label{tab:summary_sag}
\end{table}

\begin{figure*}[thb]
    \centering
    \includegraphics[width=1\textwidth]{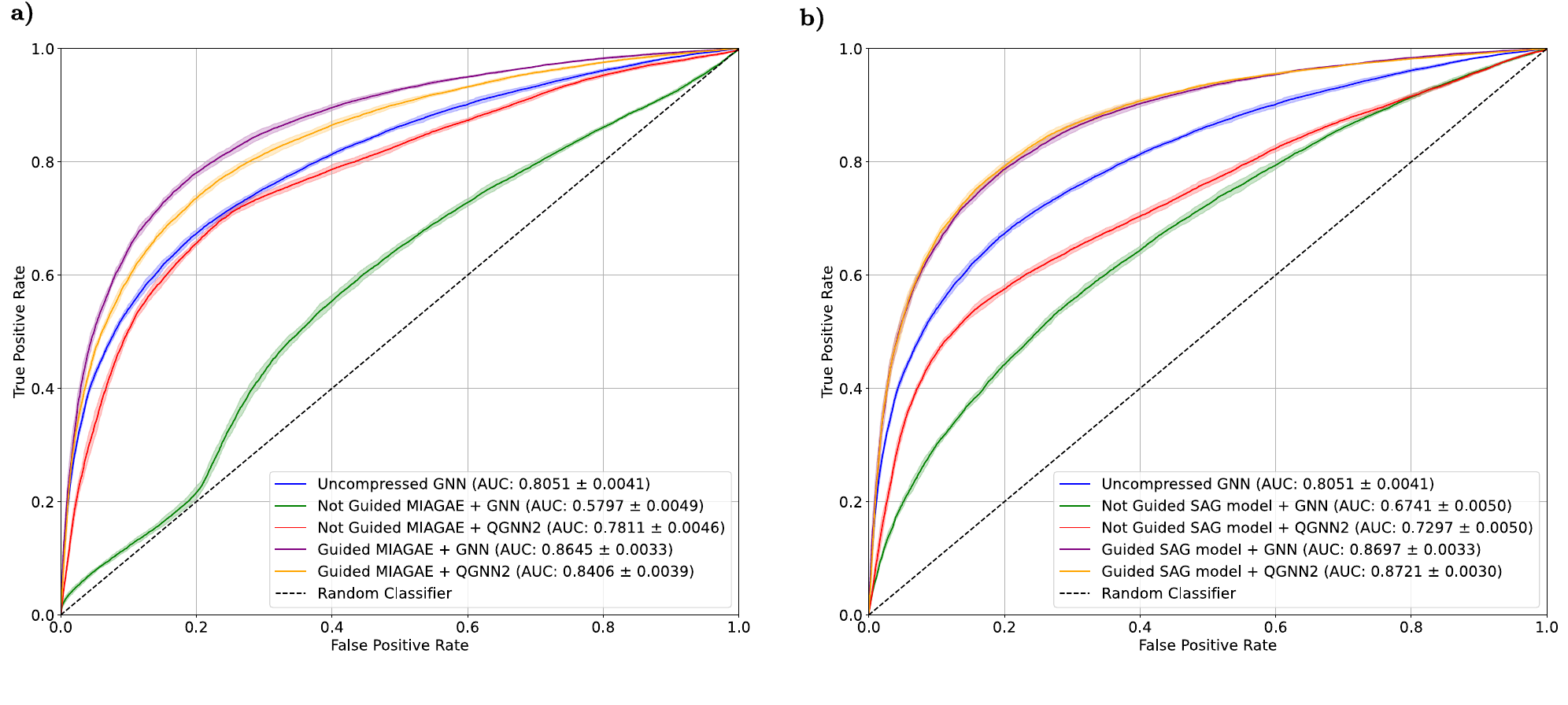}
    \caption{Receiver operating characteristic (ROC) curves comparing the three training paradigms. a) Results using the MIAGAE autoencoder for compression for Classical GNN and QGNN2 classifiers. b) Results using the SAG model autoencoder for compression for Classical GNN and QGNN2 classifiers.
A random classifier and the Uncompressed GNN classifier are included as baselines.}
    \label{fig:plot1}
\end{figure*}

The dataset is subsampled and split into 50,000 samples for training, 5,000 samples for validation and 50,000 samples for testing. The full training set is used only when training the graph autoencode in the two-step approach. To train all the models using the GGC framework, only 10,000 samples are used from this set. The test set is divided into 5 folds and the models are tested on each fold separately. Through this resampling we obtain an estimate of the mean and the standard deviation of the performance measures. The details of the evaluation methodology including the optimization of hyperparameters can be found in Appendix \ref{appendix:hyperparameter}.

We evaluate the performance of the three training paradigms using the Receiver Operating Characteristic (ROC) curves, which are depicted in Figure~\ref{fig:plot1}. Furthermore, the test ROC-AUC scores of the MIAGAE and SAG model autoencoders are presented in Tables \ref{tab:summary_miagae} and \ref{tab:summary_sag}, respectively. The GGC framework outperforms the other two training paradigms.
Specifically, training the graph autoencoders as a separate preprocessing step degrades the performance of the classification models and it is even worse than the GNN baseline using the uncompressed data. Overall, the best performing model is the combination of the SAG model autoencoder with the QGGN2 classifier using the GGC framework.


\section{Conclusion}

In this work, we introduce the Guided Graph Compression (GGC) framework, which enables processing high-dimensional graphs in quantum machine learning models. The GGC framework represents the first application of graph autoencoders for dimensionality reduction in quantum machine learning, providing an automated approach to compressing graph datasets without losing information critical to graph classification tasks. Furthermore, it constitutes a new method for learning graph representations that balance between expressivity and the computational resources required to analyze them.

We demonstrate the effectiveness of the proposed framework on a real-world application of fundamental importance: the Jet Tagging task, a critical graph classification problem in high-energy physics collider experiments. Our numerical results show that GGC outperforms both classical baseline models and the conventional two-step approach of separately training an autoencoder and quantum classifier.

While our baseline employs a classical GNN classifier, we acknowledge that other classical state-of-the-art approaches have reported higher ROC-AUC scores~\cite{qu2022particle, qu2020jet}, often leveraging significantly larger model architectures and training datasets. Such resource-intensive methods remain impractical for current quantum simulations and hardware, underscoring the relevance of our work in practical settings where computation is constrained.

Although the best result in our study was obtained using a quantum classifier, we do not claim a general quantum advantage. Rather, our contribution lies in introducing a unified framework for graph compression and classification that is compatible with both classical and quantum machine learning paradigms. This framework enables efficient handling of reduced-size datasets, making it especially suitable for hybrid or near-term quantum-classical applications.

Future work will focus on deploying this framework on real quantum devices and exploring its scalability and performance under varying graph sizes and model complexities. We hope that this direction encourages further research into graph-based learning tasks tailored to the limitations and opportunities of emerging quantum technologies.

\section*{Acknowledgements} 
SV and MG are supported by CERN through the CERN Quantum Technology Initiative. VB is supported by an ETH Research Grant (grant no.~ETH C-04 21-2). EA is supported by the European Commission (QUADRATURE: HORIZON-EIC-2022-PATHFINDEROPEN-01-101099697) and Generalitat de Catalunya ICREA Academia Award 2024. EFC is supported by grant PID2023-146520OB-C22, funded by MICIU/AEI/10.13039/501100011033, and by the Ministry for Digital Transformation and of Civil Service of the Spanish Government through the QUANTUM ENIA project call – Quantum Spain project, and by the European Union through the Recovery, Transformation and Resilience Plan – NextGenerationEU within the framework of the Digital Spain 2026 Agenda.

\section*{Data availability}
The dataset used for this work is publicly available on Zenodo \cite{komiske_2019_3164691}.

\section*{Code availability}
The code used to perform the experiments in this work is publicly available in the following GitHub repository: \url{https://github.com/mikelcasals/GGC_4_QGNNs}. PyTorch \cite{paszke2019pytorch}, PyTorch Geometric \cite{fey2019fast} and PennyLane \cite{bergholm2018pennylane} have been used for the implementation of the framework.

\bibliography{references}

\appendix
\onecolumngrid

\section{Feature Engineering of the Jet Tagging Dataset}
\label{appendix:feat_engineering}

The raw Pythia8 Quark and Gluon Jets for Energy Flow dataset \cite{komiske2019energy} is composed of 2 million jets, distributed in a balanced way between the two classes. Each jet is composed of a variable number of particles and each particle is characterized by 4 features summarized in Table \ref{tab:particle_features}. 


\begin{table*}[h]
    \centering
    \begin{tabular}{@{}ll@{}}
        \toprule
        \textbf{Feature}    & \textbf{Description}                                             \\ \midrule
        $p_T$               & Transverse momentum of the particle                              \\
        $y$                 & Rapidity of the particle                                         \\
        $\phi$              & Azimuthal angle of the particle                                  \\
        \texttt{pdgid}      & Particle Data Group ID, identifying the type of particle         \\ \bottomrule
    \end{tabular}
    \caption{Summary of particle features in the jet dataset.}
    \label{tab:particle_features}
\end{table*}

As it is done in the literature \cite{qu2022particle}, these features are augmented to the ones described in Table \ref{tab:variables}. 
\begin{table}[H]
    \centering
    \begin{tabular}{llp{10cm}}
        \toprule
        \textbf{Category} & \textbf{Variable} & \textbf{Definition} \\ 
        \midrule
        \multirow{7}{*}{Kinematics} 
        & $\Delta \eta$ & difference between the particle and the jet axis in pseudorapidity $\eta$ \\ 
        & $\Delta \phi$ & difference between the particle and the jet axis in azimuthal angle $\phi$ \\ 
        & $\log p_T$ & logarithm of the particle's transverse momentum $p_T$ \\ 
        & $\log E$ & logarithm of the particle's energy E \\ 
        & $\log \frac{p_T}{p_{T(\text{jet})}}$ & logarithm of the particle's transverse momentum $p_T$ relative to the total jet transverse momentum $p_T$ \\ 
        & $\log \frac{E}{E(\text{jet})}$ & logarithm of the particle's energy relative to the total jet energy \\ 
        & $\Delta R$ & angular separation between the particle and the jet axis $\left( \sqrt{(\Delta \eta)^2 + (\Delta \phi)^2} \right)$ \\ 
        \midrule
        \multirow{6}{*}{\begin{tabular}{c}Particle\\identification\end{tabular}} 
        & \texttt{charge} & electric charge of the particle \\ 
        & \texttt{Electron} & if the particle is an electron (\texttt{|pid|==11}) \\ 
        & \texttt{Muon} & if the particle is a muon (\texttt{|pid|==13}) \\ 
        & \texttt{Photon} & if the particle is a photon (\texttt{|pid|==22}) \\ 
        & \texttt{CH} & if the particle is a charged hadron (\texttt{(|pid|==211) + (|pid|==321)*0.5 + (|pid|==2212)*0.2}) \\ 
        & \texttt{NH} & if the particle is a neutral hadron (\texttt{(|pid|==130) + (|pid|==2112)*0.2}) \\ 
        \bottomrule
    \end{tabular}
    \caption{Augmented features and their definitions.}
    \label{tab:variables}
\end{table}

The following features are normalized by dividing them by their maximum absolute value in the training set: $\log p_T$, $\log E$, $\log \frac{p_T}{p_{T(\text{jet})}}$, $\log \frac{E}{E(\text{jet})}$ and $\Delta R$.

To characterize the jets as fully connected graphs, the particles are represented as nodes and the edge weight between two nodes is calculated as the distance between the respective particles in the pseudorapdity and azimuthal angle space, that is,

\[d_{AB} = \sqrt{(\Delta\eta_A -\Delta\eta_B)^2 + (\Delta\phi_A-\Delta\phi_B)^2}\].

This is similar to what it is done in \cite{qu2020jet}.

\section{Evaluation Methodology and Hyperparameter Optimization}
\label{appendix:hyperparameter}
\begin{figure}[H]
    \centering
    \includegraphics[width=0.75\textwidth]{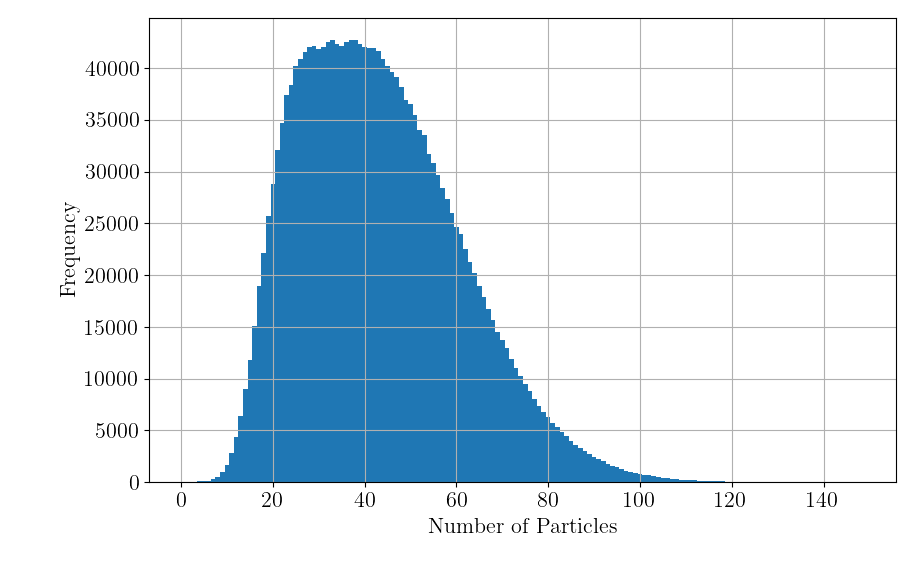}
    \caption{Distribution of the number of particles per jet.}
    \label{fig:particle_distribution}
\end{figure}

The dataset used in this work is subsampled and split into 50k samples for training, 5k samples for validation and 50k samples for testing. The subsampling is carried out preserving the distribution of the number of particles per jet in the original dataset, which can be seen in Figure \ref{fig:particle_distribution}. 

Due to hardware constraints, the 50k samples for training are only used completely when training the autoencoders as a separate preprocessing step. To train the rest of the models, only 10k samples are used from the training set. 

Every model is trained for 100 epochs, with an early stopping limit of 25 epochs with no improvement on the validation loss.

For the graph autoencoders, the following hyperparameters are fixed for all the training paradigms.
\begin{itemize}
    \item Depth: 3
    \item Shapes: 13, 13, 2
    \item Compression rate: 0.4
    \item Kernels: 1
\end{itemize}

An optimization of these hyperparameters was not done since the search space would be too large. The compression rate was chosen to be 0.4 because, with an encoder depth of 3, graphs are compressed to 6.4\% of their original size. This means that, for the dataset used, the compressed graphs contain at most 10 nodes. Therefore, only quantum circuits with at most 10 qubits need to be simulated when training the quantum classifiers.

The hyperparameters that have been optimized along with their respective possible values are the following.

\begin{itemize}
    \item Batch size: 32, 64, 128, 256, 512, 1024, 2048
    \item Learning rate: 0.001, 0.01, 0.1
    \item Number of layers (for the quantum classifiers): 2, 4, 6
    \item Classification weight $\lambda$ (for the GGC framework): 0, 0.1, 0.2, 0.3, 0.4, 0.5, 0.6, 0.7, 0.8, 0.9, 1
\end{itemize}

To optimize the hyperparameters for classical models, an exhaustive grid search is conducted over batch size and learning rate. The classification weight is initially fixed to 0.5. Once the optimal values of batch size and learning rate are found, the classification weight is optimized. For quantum classifiers, however, a sequential grid search is used to manage the higher training time. In this sequential approach, the optimization of each hyperparameter is done one by one, using the best values found for previously optimized hyperparameters. The initial default values are: batch size of 32, learning rate of 0.001, 2 layers, and a classification weight of 0.5. We begin by optimizing the batch size, keeping the other hyperparameters fixed. Once the batch size is tuned, we proceed to optimize the learning rate, followed by the number of layers, and finally, the classification weight.

The criteria to find the optimal set of hyperparameters is to choose the models with the best Receiving Operating Characteristic - Area Under the Curve (ROC-AUC) score. Then, the overall best models are tested on the test set using the 5-fold approach to obtain the mean and standard deviation.

To ensure reproducibility in the results, the random seed is fixed to 42. All the models have been trained on an Intel(R) Xeon(R) CPU E5-1630 v3 CPU at 3.70GHz with 8 cores and 32 GB of RAM.

The optimal hyperparameters found for each of the evaluated training paradigms can be found in Tables \ref{tab:hp_uncompressed}, \ref{tab:hp_gaes}, \ref{tab:hp_notguided_miagae}, \ref{tab:hp_notguided_sag_model}, \ref{tab:hp_guided_miagae} and \ref{tab:hp_guided_sag_model}.

\begin{table}[H]
    \centering
    \begin{tabular}{lcc}
        \toprule
        & \textbf{Batch} & \textbf{Learning} \\ 
        &\textbf{Size} & \textbf{Rate} \\
        \midrule
        Uncompressed Classical GNN & 32 & 0.1 \\
        \bottomrule
    \end{tabular}
    \caption{Optimal hyperparameters found for the Uncompressed Classical training paradigm.}
    \label{tab:hp_uncompressed}
\end{table}

\begin{table}[H]
    \centering
    \begin{tabular}{lcc}
        \toprule
        & \textbf{Batch} & \textbf{Learning} \\ 
        &\textbf{Size} & \textbf{Rate} \\
        \midrule
        SAG model & 1024 & 0.001 \\
        MIAGAE & 128 & 0.01 \\
        \bottomrule
    \end{tabular}
    \caption{Optimal hyperparameters found for the pretrained graph autoencoders used for the Not Guided Graph Compression training paradigm.}
    \label{tab:hp_gaes}
\end{table}

\begin{table}[H]
    \centering
    \begin{tabular}{lccc}
        \toprule
        & \textbf{Batch} & \textbf{Learning} & \textbf{Number of}\\ 
        &\textbf{Size} & \textbf{Rate} & \textbf{Layers}\\
        \midrule
        MIAGAE + GNN & 2048 & 0.001 & - \\
        MIAGAE + QGNN1 & 2048 & 0.001 & 2\\
        MIAGAE + QGNN2 & 32 & 0.001 & 6\\
        \bottomrule
    \end{tabular}
    \caption{Optimal hyperparameters found for the Not Guided Graph Compression training paradigm using the MIAGAE graph auteoncoder.}
    \label{tab:hp_notguided_miagae}
\end{table}

\begin{table}[H]
    \centering
    \begin{tabular}{lccc}
        \toprule
        & \textbf{Batch} & \textbf{Learning} & \textbf{Number of}\\ 
        &\textbf{Size} & \textbf{Rate} & \textbf{Layers}\\
        \midrule
        SAG model + GNN & 32 & 0.1 & - \\
        SAG model + QGNN1 & 32 & 0.001 & 4\\
        SAG model + QGNN2 & 32 & 0.1 & 6\\
        \bottomrule
    \end{tabular}
    \caption{Optimal hyperparameters found for the Not Guided Graph Compression training paradigm using the SAG model graph auteoncoder.}
    \label{tab:hp_notguided_sag_model}
\end{table}

\begin{table}[H]
    \centering
    \begin{tabular}{lcccc}
        \toprule
        & \textbf{Batch} & \textbf{Learning} & \textbf{Number of} & \textbf{Classification Weight} \\ 
        &\textbf{Size} & \textbf{Rate} & \textbf{Layers} &\textbf{$\lambda$}\\
        \midrule
        MIAGAE + GNN & 32 & 0.01 & - & 0.9\\
        MIAGAE + QGNN1 & 128 & 0.01 & 4 & 0.8\\
        MIAGAE + QGNN2 & 32 & 0.001 & 2 & 0.5\\
        \bottomrule
    \end{tabular}
    \caption{Optimal hyperparameters found for the Guided Graph Compression training paradigm using the MIAGAE graph auteoncoder.}
    \label{tab:hp_guided_miagae}
\end{table}

\begin{table}[H]
    \centering
    \begin{tabular}{lcccc}
        \toprule
        & \textbf{Batch} & \textbf{Learning} & \textbf{Number of} & \textbf{Classification Weight} \\ 
        &\textbf{Size} & \textbf{Rate} & \textbf{Layers} &\textbf{$\lambda$}\\
        \midrule
        SAG model + GNN & 256 & 0.01 & - & 0.8\\
        SAG model + QGNN1 & 32 & 0.001 & 6 & 0.5\\
        SAG model + QGNN2 & 32 & 0.001 & 6 & 0.8\\
        \bottomrule
    \end{tabular}
    \caption{Optimal hyperparameters found for the Guided Graph Compression training paradigm using the SAG model graph auteoncoder.}
    \label{tab:hp_guided_sag_model}
\end{table}

\end{document}